# Web Page Categorization Using Artificial Neural Networks

S. M. Kamruzzaman

Department of Computer Science and Engineering
Manarat International University, Dhaka, Bangladesh
E-mail: smzaman@gmail.com, smk.cse@manarat.ac.bd

**ABSTRACT**

**Web page categorization is one of the challenging tasks in the world of ever increasing web technologies. There are many ways of categorization of web pages based on different approach and features. This paper proposes a new dimension in the way of categorization of web pages using artificial neural network (ANN) through extracting the features automatically. Here eight major categories of web pages have been selected for categorization; these are business & economy, education, government, entertainment, sports, news & media, job search, and science. The whole process of the proposed system is done in three successive stages. In the first stage, the features are automatically extracted through analyzing the source of the web pages. The second stage includes fixing the input values of the neural network; all the values remain between 0 and 1. The variations in those values affect the output. Finally the third stage determines the class of a certain web page out of eight predefined classes. This stage is done using back propagation algorithm of artificial neural network. The proposed concept will facilitate web mining, retrievals of information from the web and also the search engines.**

## 1. INTRODUCTION

The rapid growth nature of world wide web (www) necessitates providing newly invented techniques for automatic categorization of the web pages into different classes or categories [7] [13]. This is necessary to assist the end users as well as the web servers and search engines to find the desired web sites [11] [12]. Automatic categorization of web pages has been studied extensively, and most of these categorization techniques are usually based on similarity between documents contents or their structures [1] [2] [3].

Most of the web page categorization techniques focus on certain features of web pages, which are limited [4] [5]. There are some categorization techniques that classify the pages using the meta keyword, hyperlinks structures, document structure, and automatic text categorizations [4] [8] [14]. Web pages are designed for human; anyone can easily classify a page when he/she goes through the page [15]. But this is quiet boring and time consuming to search a page of intended class [6].

A technique for web page categorization using artificial neural network (ANN) through automatic feature extraction is proposed. The main objective behind this task is to provide an efficient way for categorization of web pages. This will facilitate the different search engines to classify the web pages more efficiently and also to provide a rich web directory. The end user will also be facilitated to find the web page of their desired classes. With the rapid growth of the www, there is an increasing need to provide automated assistance to web users for web page categorization [9] [10]. As we have observed that a users fails to find out their intended web site if the pages are classified only on limited features and similarities. So a new and efficient technique is proposed to facilitate the user for categorization the web pages. This work covering eight major classes of web pages including business & economy, education, government, entertainment, sports, news & media, job search, and science. The web pages are classified based on the five different features. The whole process is done by three successive stages of operations.

## 2. THE PROPOSED APPROACH

There are several categorization approaches defined by various researchers in different courses of time to face the demand of users. Because of rapid growth of web pages per day, an efficient technique is proposed here to classify the web pages based on the five features extracted from a page and the categorization is done in three successive stages. Through analysis of about 500 web pages it is found that all the web developers and the designers try to express the motto and the theme of the organization. The theme is expressed by the total structure of the home page. The intention of the designer is always to make the visitor spent more time in his site. So he tries to design the home page with extra care and build the home page structure to fulfill the intention. So the five features are picked up that make the site different from other types. The features are home page structure, which is the ratio of internal and external links, amount of dynamic/static pages, frequency of images, availability of animations and the predefined buzzwords. In this proposed approach, eight major classes are selected from different web directory. The neural network are trained and tested by using those classes. The proposed approach is done through the following stages:
1) Automatic features extraction through analyzing the home page source.
2) Fixing the values for the input nodes of the networks.
3) Classifying web pages by the neural networks.

### 2.1 Features Extraction

In the proposed approach features are extracted by analyzing the source of a web page. By going through the tags of the source file the features can easily get.

### 2.1.1 Ratio of Internal & External Links

The site structure is defined by the internal or external links used in the page. By analyzing the reference or information pages (science/education/job site) more external links are found than the commercial web sites (business and economy, news and media, sports sites).

### 2.1.2 Frequency of Buzzwords

The proposed approach defines some buzzwords that make a site to be a certain class. The frequency of buzzwords from same class increases the probability of that site to be of that class. The values of input layer of ANN are determined by calculating the buzzwords.

### 2.1.3 Number of Images

This is one of the major considerations that used images in a web site reflect the theme of home page. In the proposed approach it is not concern to image categorization methods but of finding the area covered in the home page by the images. More number of images proves that the web page is more colorful.

**Table 1:** Selected buzzwords.

| Classes | Buzz Words |
|---|---|
| Business, Economy | Business, Trade, Investment, Credit, Cash, Trade, Commerce, Loan, Support, Product, Service, Offer. |
| Education | Career, Student, Faculty, Degree, Graduate, Education, Research, Admission, Prospects. |
| Government | Policy, Ministry, President, Government, Activity. |
| News, Media | News, Media, Editor, Culture, Archives, Latest, Update, Current, Affairs. |
| Entertainment | Music, Entertainment, Dating, Fun, Love, Artist, Free, Match, Friendship. |
| Science | Science, Research, Technology. |
| Sports | Team, Sports, Matches, Schedule, Scores. |
| Job Site | Career, Experience, Job, Seek, Vacancy, Resume, Application, Location, Employment, Offer. |

### 2.1.4 Availability of Animation

Animations are mostly used in the business sites and advertising sites. A company logo is animated and a number of logos are used in the job sites. Flash animations, script animations fall into this. If we visit a web site of a well-known university we find that animations are rarely used there. Science sites also fall into this category.

### 2.1.5 Static Vs Dynamic Pages

The number of static or dynamic pages used in a site also gives a clue to classify the site. It is investigated that the news and media, sports sites are most frequently updated. The news pages are updated daily and there are some sites those are updated in every five seconds. Again the business or information pages are rarely updated. So the more number of dynamic pages increases the probability to be news site or job site or sports site whether the fewer number of dynamic pages indicates that the page is either a science or a government or education site.

### 2.2 Fixing the Values for ANN

Extracted features are setting as input of the ANN. The site structure provides the first input value. This is found by dividing the number of hyperlinks to out side of the domain by the number of hyperlinks to inside the domain. It is seen that a business web site rarely provides any link to out side of the domain. They always try to focus themselves and try to keep the visitors stay long in their site whether a job site provides more links to out side of their domain as they can be said as an advertising site. They provide the links to the web sites of various companies. A fun or entertainment site also provides links to their sponsors.

Buzzwords counted by analyzing the frequency of certain words in the home page of a web site. This gives us the use of frequency of keywords used in the home page. Several key words are selected for desired categories, which are searched in the body of the page and the most frequent keyword, and its value gives us another input of the ANN.

Crowd of images means the frequency of image used in a page. If we visit the standard pages of various classes we see an educational institute use less colors and images than a job site or a news/sports site. The probability of use of images in a science and engineering or a personal home page is very low. The more image used gives the impression that the page is more colorful. It is generally known that the information/research pages are less colorful than a commercial page. The use of Image is also lower.

Availability of animations means how much area is covered by animations in the home page of a web site. Basically animations are mostly used to attract the attention of a visitor to the product of some manufacturing companies, which fall into the business and economy category. A university website generally consists of fewer animations than a web site of a paint manufacturing company.

Web page type provides a value for the input layer of ANN. The more dynamic pages used in a site means the more possibility to be a news site or sports site. And the less use of dynamic page means the possibility to be a science and engineering site. This value remains from 0 to 1 depending on the percentage of dynamic pages. Table 2 shows the input values to ANN for dynamic pages.

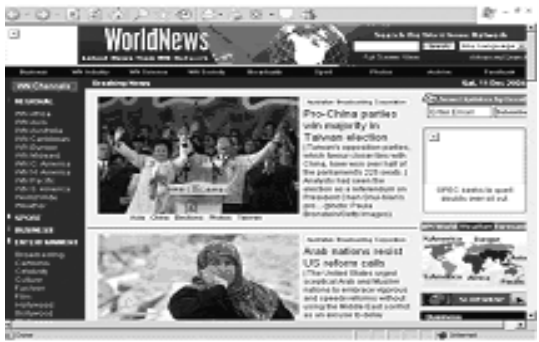

**Fig. 1:** A complete dynamic web site.

### 2.3 Proposed Network Architecture

For the network we used a 5-5-3 architecture. The input vector of this network consists of 5 elements where each neuron represents one element. In this architecture one hidden layer with 5 neurons are used. Output of the network consists of 3 neurons, to show the output pattern based on our eight classes. By three neurons we can classify the web pages into eight categories. There are lots of possibilities to be fraction value to come as output. To overcome this situation we are converting the values from 0.0 to 0.49 into 0 and the values from 0.50 to 1.0 into 1. The targeted classes in binary manner are shown in Table 3.

**Table 2:** Input values for dynamic pages.

| % of Dynamic pages | Input into ANN |
|---|---|
| 90% - 100% | 1.0 |
| 80% - 90% | 0.9 |
| 70% - 80% | 0.8 |
| 60% - 70% | 0.7 |
| 50% - 60% | 0.6 |
| 40% - 50% | 0.5 |
| 30% - 40% | 0.4 |
| 20% - 30% | 0.3 |
| 10% - 20% | 0.2 |
| 0% - 10% | 0.1 |
| No dynamic page | 0.0 |

**Table 3:** Output patterns.

| Classes of the web pages | Output Pattern |
|---|---|
| Business and Economy | 0 0 0 |
| Education | 0 0 1 |
| Government | 0 1 0 |
| News and Media | 0 1 1 |
| Sports | 1 0 0 |
| Job Search | 1 0 1 |
| Entertainment | 1 1 0 |
| Science | 1 1 1 |

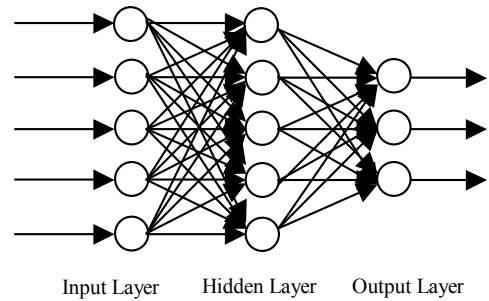

Input Layer    Hidden Layer    Output Layer

**Fig. 2:** Architecture of the proposed ANN.

### 2.4 Training and Testing Data

A good number of training examples needed to get efficient response from the system. But this is really hard to get available training example. For this purpose 500 of home pages of different web sites are collected. All the samples were split into two groups: the training set and the testing set. The training set was comprised of 40% of the total samples and the rest of the samples are used for testing the system.

## 2.5 Network Testing & Performance

Trained networks needed to be tested to measure the performance of the network. The evolution of the true performance of any system depends on the organization of testing data set. The testing data set should be enough strong to reflect the real world situation. For this, the testing data set demands various types of input pattern that may be arise in the real world situation. We tried to make our testing data set from different categories and also from domains of different countries.

## 3. EXPERIMENTAL RESULTS

To show the effectiveness of the proposed approach the experiment has done by testing the system in different ways and then comparing their result to the final result. The system is tested firstly using the known pages of 200 pages and then for 300 unknown pages. The following table shows the results of the system when network was tested:

**Table 4:** Experimental results.

| Types of pages | No. of pages right classified | No. of pages wrong classified |
| --- | --- | --- |
| Business & Economy | 35 | 14 |
| Education | 23 | 6 |
| Government | 12 | 8 |
| News & Media | 29 | 5 |
| Sports | 18 | 7 |
| Entertainment | 31 | 18 |
| Job Search | 26 | 11 |
| Science | 38 | 19 |
| **Total** | **212** | **88** |

## 4. CONCLUSION

Automated categorization of web pages can lead to better web retrieval tools with the added convenience of selecting among properly organized directories. In this paper a theme based web page categorization is proposed which extract the features automatically through analyzing the html source, and categorize the web pages into eight major classes using back propagation algorithm. The web pages are categorized based on five major characteristics and similarities of different pages of same types.

## REFERENCES

[1] Aijun An and Xiangji Huang, "Feature selection with rough sets for web page categorization", York University, Toronto, Ontario, Canada.

[2] Arul Prakash Asirvhatam and Kranti Kumar Ravi." Web Page Categorization based on Document Structure", International Institute of Information Technology, Hyderabad, India 500019.

[3] Chandra Chekuri & Michale,Stamford University; Prabhakar Gold wasser and Eli Uphal, "Web Search Using Automatic Categorization", IBM alamden Research Center, 650 Harry Road,San Jose CA 95120.

[4] D. Boley, M. Gini, R. Gross, E-H. S. Han, K. Hastings, G. Karypis, V. Kumar, B. Mobasher, and J. Moore." Partitioning-based clustering for web document categorization", Decision Support System, 1999.

[5] Dou Shen Zheng Chen and Yu chang Lu, "Web page categorization through summarization", Tsinghua University.

[6] E.J.Glover, G.W.F.S. Lawrence, W.P.Birmingham, A.Kruger, C.L.Giles, and D.M.Pennock. "Improving category specific web search by learning query modifications". In symposium on applications and the internet, SAINT 2001, san diego, California, January 8-12, 2001.

[7] Gerry McGovern. "A step to step a appoarch to web page categorization". www.gerrymcgovern.com.

[8] H.J.Oh, S.H.Myaeng, and M.H.Lee, "A practical hypertext categorization method using links and incrementally available class information", In *SIGIR 2000*, Athens, Greece, 2000.

[9] H.Yu, J.Han, and K.C.C.Chang.Pebl, "Positive-example based learning for web page categorization using SVM", In KDD, Edmonton, Alberta, Canada, 2002.

[10] Hui Yang & Tat-Seng Chua " Effectiveness of web page categorization on Finding List Answer ", National University of Singapore.

[11] Hwanjo Yu,kevin chen chuan chang,Jiawei Han, "Heterogeneous learner for web page categorization", University of Illinois at Urbana-Champaign.

[12] J.Hayes and W.S.P." A system for content-based indexing of a database of news stories". In Proceedings of Second Annual Conference on Innovatative Applications of Artificial Intelligence, pages 1–5, 1990.

[13] J.Yi and N.Sudershesan, "A classifier for semi-structured documents", In KDD 2000, Boston, MA USA, 2000.

[14] John M. Pierre, "On the Automated Categorization of Web Sites", Linkoping University Electronic Press Linkoping, Sweden.

[15] K.Matsuda and T.Fukushima."Task-oriented world wide web retrieval by document type categorization", In *CIKM '99*, Kansas City, Mo, USA.